\documentclass{article}

\usepackage[preprint,nonatbib]{neurips_2023}

\usepackage[utf8]{inputenc} 
\usepackage[T1]{fontenc}    
\usepackage{url}            
\usepackage{booktabs}       
\usepackage{amsfonts}       
\usepackage{nicefrac}       
\usepackage{microtype}      
\usepackage{xcolor}         
\usepackage{amsmath}
\usepackage{graphicx}
\usepackage{multirow}
\usepackage{booktabs}
\usepackage{wrapfig}
\usepackage{enumitem}
\usepackage[pagebackref,breaklinks,colorlinks]{hyperref}

\usepackage[capitalize]{cleveref}
\crefname{section}{Sec.}{Secs.}
\Crefname{section}{Section}{Sections}
\Crefname{table}{Table}{Tables}
\crefname{table}{Tab.}{Tabs.}

\title{LayerDiffusion: Layered Controlled Image Editing with Diffusion Models}

\author{%
  \vspace{0.4em}
  Pengzhi Li \textsuperscript{\rm 1}\quad
  Qinxuan Huang \textsuperscript{\rm 1}\quad
  Yikang Ding \textsuperscript{\rm 1} \quad
  Zhiheng Li \textsuperscript{\rm 1}\thanks{Corresponding author.} \quad
   \vspace{0.6em}
  \\
  \textsuperscript{\rm 1} Tsinghua University   \quad
  \\
}
\begin{document}

\maketitle
\begin{abstract}
Text-guided image editing has recently experienced rapid development. However, simultaneously performing multiple editing actions on a single image, such as background replacement and specific subject attribute changes, while maintaining consistency between the subject and the background remains challenging. In this paper, we propose \textit{LayerDiffusion}, a semantic-based layered controlled image editing method. Our method enables non-rigid editing and attribute modification of specific subjects while preserving their unique characteristics and seamlessly integrating them into new backgrounds. We leverage a large-scale text-to-image model and employ a layered controlled optimization strategy combined with layered diffusion training. During the diffusion process, an iterative guidance strategy is used to generate a final image that aligns with the textual description. Experimental results demonstrate the effectiveness of our method in generating highly coherent images that closely align with the given textual description. The edited images maintain a high similarity to the features of the input image and surpass the performance of current leading image editing methods. 
\textit{LayerDiffusion} opens up new possibilities for controllable image editing.

\end{abstract}

\section{Introduction}

Given a single image of your pet, it can be imagined embarking on a worldwide journey and performing specific actions in any location. Generating such an image is a challenging and fascinating task in image editing. It entails preserving the specific subject's unique characteristics in new backgrounds and ensuring their seamless integration into the scene, harmoniously and naturally, while simultaneously accommodating multiple editing actions.

Recently, significant progress has been made in the development of deep learning-based large-scale text-to-image models~\cite{rombach2022high,saharia2022imagen,ramesh2022hierarchical}. These models can generate high-quality synthetic images based on text prompts, enabling text-guided image editing and producing impressive results. As a result, numerous text-based image editing methods~\cite{tumanyan2022plug,hertz2022prompt,gal2022textual_in,chefer2023attend,ruiz2022dreambooth,couairon2022diffedit,su2022dual} have emerged and evolved. However, such models cannot mimic specific subject characteristics. Even with the most detailed textual descriptions of an object, they may generate instances with different appearances and still struggle to maintain background consistency. Thus, the current leading image editing methods encounter several challenges, including rigid editing limited to specific domain images~\cite{patashnik2021styleclip,hertz2022prompt}, the inability to simultaneously edit both the background and specific subjects, and the requirement for additional auxiliary input information~\cite{ruiz2022dreambooth,nichol2021glide,avrahami2022spatext,bar2023multidiffusion}. These issues hinder the advancement of controllable image editing. 

In this paper, we propose a semantic-based layered controlled image editing method, which we call \textit{LayerDiffusion}, to alleviate these issues. By simply inputting textual descriptions of multiple editing actions, along with the target image and a reference image, we can perform non-rigid editing and attribute modification of specific subjects, generating images consistent with the textual descriptions while maintaining the consistency of the specific subject and background features with the input image. As shown in~\cref{fig:teaser}, we can make a dog jump in a forest or a giraffe lies on a beach or modify their shapes and attributes in the original scene. 

To implement our method, we leverage the robust and high-quality image generation capabilities of a large-scale text-to-image model~\cite{rombach2022high}. 
Our method comprises a well-defined sequence of steps.
Initially, we utilize a mask to eliminate interference from foreground objects effectively. Subsequently, we apply a layered controlled optimization strategy to optimize the text embeddings acquired from the text encoders~\cite{raffel2020exploring}, following the segmentation of the target text. This process aims to generate image backgrounds that exhibit a remarkable similarity to the reference images.
Next, we employ a layered diffusion training strategy to fine-tune the model, thereby augmenting its ability to preserve the similarity between the specific subjects, backgrounds, and input images.
Finally, during the diffusion process with the fine-tuned model, we adopt an iterative guidance strategy, where a highly constrained text embedding is iteratively employed to denoise the images. Consequently, this generates a final image aligning with the textual description.

We emphasize the contributions of each component in our method through ablation studies and compare our approach with other relevant image editing methods~\cite{kawar2022imagic,tumanyan2022plug,meng2021sdedit}, clearly demonstrating superior editing quality. Furthermore, we conduct a user study to subjectively evaluate the quality of the images generated by our method, which aligns most closely with human perception. We summarize our main contributions as follows:

\begin{figure*}{}
    \centering
    \includegraphics[width=1.0\textwidth]{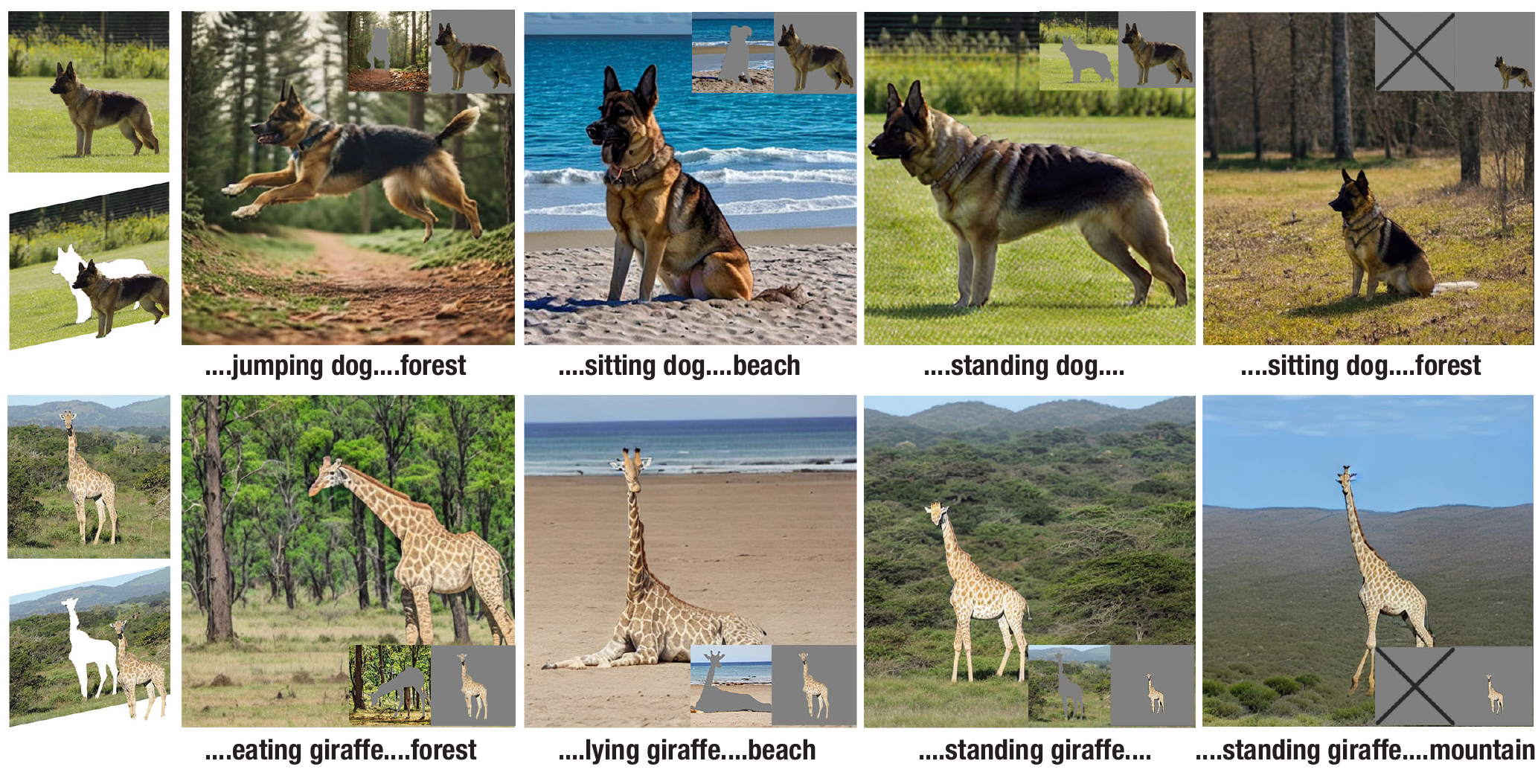}
    \caption{Our method achieves layered image editing through text descriptions, enabling simultaneous modifications of backgrounds and specific subjects, such as background replacement, object resizing, and complex non-rigid changes.}
    \label{fig:teaser}
\end{figure*}

\begin{itemize}
\item {\verb||}We propose \textit{LayerDiffusion}. To the best of our knowledge, this is the first image editing method that enables simultaneous editing of specific subjects and backgrounds using a single input image.
\item  We introduce a novel layered diffusion training framework that enables arbitrary and controllable editing of specific subjects and backgrounds.
\item Experimental results demonstrate that our method generates images with highly similar features to the input images.
\end{itemize} 

\vspace{-1em}
\section{Related Work}

Image synthesis has recently made significant advancements~\cite{abdal2022clip2stylegan,harkonen2020ganspace,patashnik2021styleclip,ramesh2021zero,zhang2021cross,abdal2021styleflow,gal2022stylegan,shen2020interpreting}. With the development of diffusion models~\cite{ho2020denoising,song2020denoising,song2019generative} in image processing tasks~\cite{kawar2022denoising,saharia2022palette,vahdat2021score,wolleb2022diffusion,blattmann2022retrieval}, new text-guided solutions have emerged in the field of image editing and produced impressive results~\cite{nichol2021glide,ramesh2022hierarchical,zhang2022sine,chefer2023attend,kawar2022imagic}. The powerful generative capabilities of diffusion models enable the generation of numerous high-quality images. Consequently, many image editing tasks~\cite{bar2022text2live,hertz2022prompt,meng2021sdedit,tumanyan2022plug}no longer require training a large-scale text-to-image model, as pre-trained models can be used for image editing based on textual descriptions. Diffusion models have tremendous potential for image editing tasks guided by text descriptions. Many studies~\cite{meng2021sdedit,kawar2022imagic,ruiz2022dreambooth,tumanyan2022plug,hertz2022prompt} have utilized pre-trained models as generative priors, which can be categorized into two approaches: training-free and fine-tuned method. SDEdit~\cite{meng2021sdedit} introduces intermediate noise to an image, which can be augmented with user-provided brush strokes, followed by denoising through a diffusion process conditioned on the desired edit. P2P~\cite{hertz2022prompt} and PnP~\cite{tumanyan2022plug} utilize cross-attention or spatial features to edit both global and local aspects of an image by directly modifying the text prompt. However, they often preserve the original layout of the source image and struggle with non-rigid transformations.

Fine-tuned methods~\cite{kawar2022imagic,ruiz2022dreambooth,kim2022diffusionclip,gal2022textual_in} have also shown remarkable performance. DiffusionCLIP~\cite{kim2022diffusionclip} leverages the CLIP~\cite{radford2021learning} model to provide gradients for producing impressive style transfer results. Textual-inversion~\cite{gal2022textual_in} and Dreambooth~\cite{ruiz2022dreambooth} fine-tune the model using multiple sets of personalized images, resulting in the synthesis of images depicting the same object in new environments.  Imagic~\cite{kawar2022imagic} fine-tunes the model by optimizing text embeddings and achieves image editing through linear interpolation of text embeddings.

Similarly, our approach leverages target text descriptions to fine-tune the model and enable various image editing operations. Dreambooth~\cite{ruiz2022dreambooth} and Imagic~\cite{kawar2022imagic} are methods that resemble our approach. However, Dreambooth requires multiple input images and often fails to produce satisfactory results when dealing with a single image. Imagic, on the other hand, faces challenges in simultaneously performing multiple editing actions, such as editing both the background and specific subjects simultaneously. In contrast, our method allows for simultaneous editing of specific subjects and the background using only a single input image.

\begin{figure*}{}
    \centering
    \includegraphics[width=1.00\textwidth]{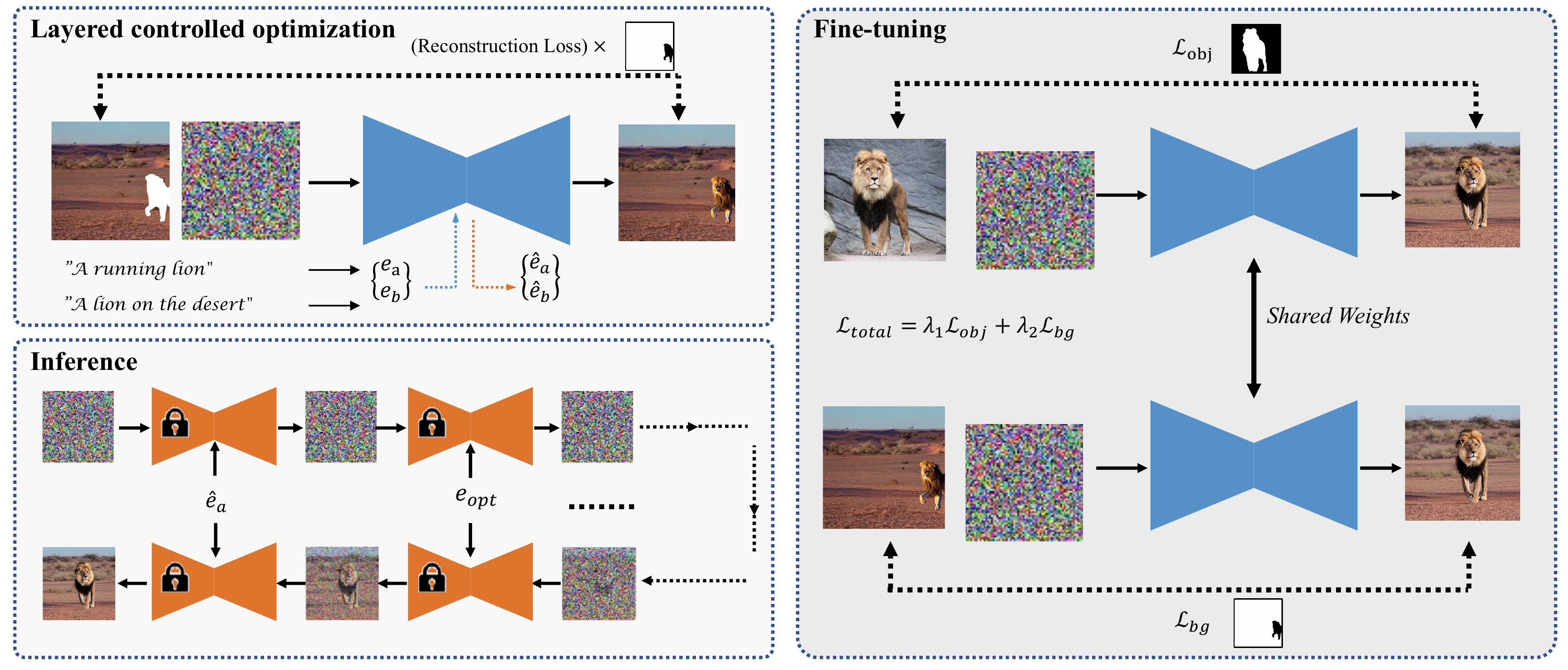}
    \caption{Our method utilizes a layered controlled optimization strategy to refine text embeddings and a layered diffusion strategy to fine-tune the diffusion model. During inference, an iterative guidance strategy is employed to directly generate images aligning with the multiple editing actions described in the input text.}
    \label{fig:pipeline}
\end{figure*}

\section{Method}

\subsection{Preliminaries}

Stable Diffusion Models (SDM) \cite{rombach2022high} is a publicly available text-to-image diffusion model trained on the LAION-5B~\cite{schuhmann2022laion} dataset.
Instead of directly operating in the image space, SDM is based on the latent diffusion method, which means the forward and reverse diffusion sampling are operated in the latent space. Given the trained autoencoder, the image $\boldsymbol{p}$ is convert to low-dimensional latent variable $\boldsymbol{x}$ at each timestep t. 
SDM also introduces an important modification in the form of text-based conditioning. During the denoising process, SDM can be conditioned on an additional input vector, which is typically a text encoding produced by a pre-trained CLIP text encoder $\mathcal{P}$. Specially, the $\mathcal{P}$ extract words from a given text prompt $\boldsymbol{y}$ and convert them into tokens, denoted by $\boldsymbol{e} = \tau_{\phi}(\boldsymbol{y})$. These tokens are further transformed into text embeddings, which are used to condition the neural network during the training process:

\begin{equation}
\min _{\theta} \mathbb{E}_{\boldsymbol{x}_{t}, \boldsymbol{x}_{0}, \boldsymbol{\epsilon} \sim \mathcal{N}(0, \boldsymbol{I})}\left[\left\|\boldsymbol{\epsilon}_{\theta}\left(\boldsymbol{x}_{t}, t, \tau_{\phi}(\boldsymbol{y})\right)-\boldsymbol{\epsilon}\right\|_{2}^{2}\right],
\label{equ: ldm}
\end{equation}

Consequently, SDM facilitates the generation of images based on textual input by employing reverse diffusion sampling in the latent space. Instead of relying on $\boldsymbol{\epsilon}_{\theta}(\boldsymbol{x}_{t}, t)$, the model utilizes a text-conditioned neural network denoted as $\boldsymbol{\epsilon}_{\theta}(\boldsymbol{x}_{t}, t, \tau_{\phi}(\boldsymbol{y}))$. We implement the proposed approach in this work by fine-tuning this pre-trained model.

\subsection{Layered Diffusion}

Our approach leverages target text descriptions to facilitate a wide range of image editing actions, including object size, property modifications, and background replacement while preserving specific subject details closely tied to the original image. To achieve this, we fine-tune a state-of-the-art diffusion model~\cite{rombach2022high}. Furthermore, we introduce a layered editing method for the background and specific foreground objects.
As illustrated in~\cref{fig:pipeline}, our method begins by separating the background. We apply a layered controlled optimization strategy to refine the segmentation text embeddings acquired from the text encoders, which come from the target text.Then we identify the optimal text embedding that aligns with the desired target background in proximity to the target text embedding. Subsequently, we employ a layered diffusion strategy to fine-tune the diffusion model. This approach enhances the model's capability to maintain similarity between specific subjects, backgrounds, and input images, allowing for finer control and precision in image editing through parameter adjustments. 
During the inference stage, we utilize an iterative guidance strategy to directly generate images that align with the multiple image editing actions described in the input text without the text embedding interpolation. Each step of the process is outlined in detail below.

\subsubsection{Layered controlled optimization}
Due to the potential interference of multiple text descriptions, optimizing text embeddings can be unstable during image editing. As a result, previous methods for image editing have often struggled to effectively modify selected object property and backgrounds simultaneously.

To this end, we aim to separate the background and foreground to reduce interference between different textual information. The target text $T$ is first fed into the Stable Diffusion model~\cite{rombach2022high} to obtain the target image $O_{t}$. 
Then $T$ is decomposed into $T_{a}$ and $T_{b}$, which describe object properties and background separately and sent to the text encoder~\cite{raffel2020exploring} to output the corresponding text embeddings  $\boldsymbol e_{a}\in \mathbb{R}^{C\times N}$ and $\boldsymbol e_{b}\in \mathbb{R}^{C\times N}$, where $C$ is the number of tokens, and $N$ is the token embedding dimension. However, $\boldsymbol{e}_{a}$ and $\boldsymbol{e}_{b}$ are in the distant embedding space, so we cannot directly perform linear interpolation on them. To make $\boldsymbol{e}_{a}$ and $\boldsymbol{e}_{b}$ match our input image background as much as possible and be in a close embedding space, we freeze the parameters of the diffusion model and optimize $\boldsymbol e_{a}$ and $\boldsymbol e_{b}$ simultaneously using the diffusion model objective~\cite{ho2020denoising}. In fact, we can optimize the initial text embedding to make it closer to the target image (modify the background) space or reference image (modify object properties) space. This process is controlled by the object mask $M$ and can be represented as follows:

\begin{equation} \label{eq_1}
\begin{bmatrix} \boldsymbol {\hat{e}}_{a}, \boldsymbol {\hat{e}}_{b} \end{bmatrix}  =\arg \min \mathbb{E}_{\boldsymbol{x}_{t}, \boldsymbol{\epsilon} \sim \mathcal{N}(0, \boldsymbol{I})}\left[ \left\| {M} *(\boldsymbol{\epsilon}- f_{\theta}\left(\boldsymbol{x}_{t}, t,\begin{bmatrix} \boldsymbol {e}_{a}, \boldsymbol {e}_{b} \end{bmatrix})\right)\right\|^{2}\right], 
\end{equation}

where $M$ is computed by Segment Anything Model (SAM)~\cite{kirillov2023sam}, and $\boldsymbol{x}_{t}$ is the noisy version of the input image, and $f_{\theta}$ means the forward diffusion process using pre-trained diffusion model. The optimized text embeddings make it meaningful to modify the linear interpolation weights of $\boldsymbol {\hat{e}}_{a}$ and $\boldsymbol {\hat{e}}_{b}$ as follows:
\begin{equation}
\boldsymbol {e}_{opt} = \alpha * \boldsymbol {\hat{e}}_{a} + (1 - \alpha) * \boldsymbol {\hat{e}}_{b},
\end{equation}
according to the experimental analysis of text embedding interpolation in Imagic~\cite{kawar2022imagic}, we tend to set the weight $\alpha$ that describes object properties to 0.7.

\begin{figure*}{}
    \centering
    \includegraphics[width=1.00\textwidth]{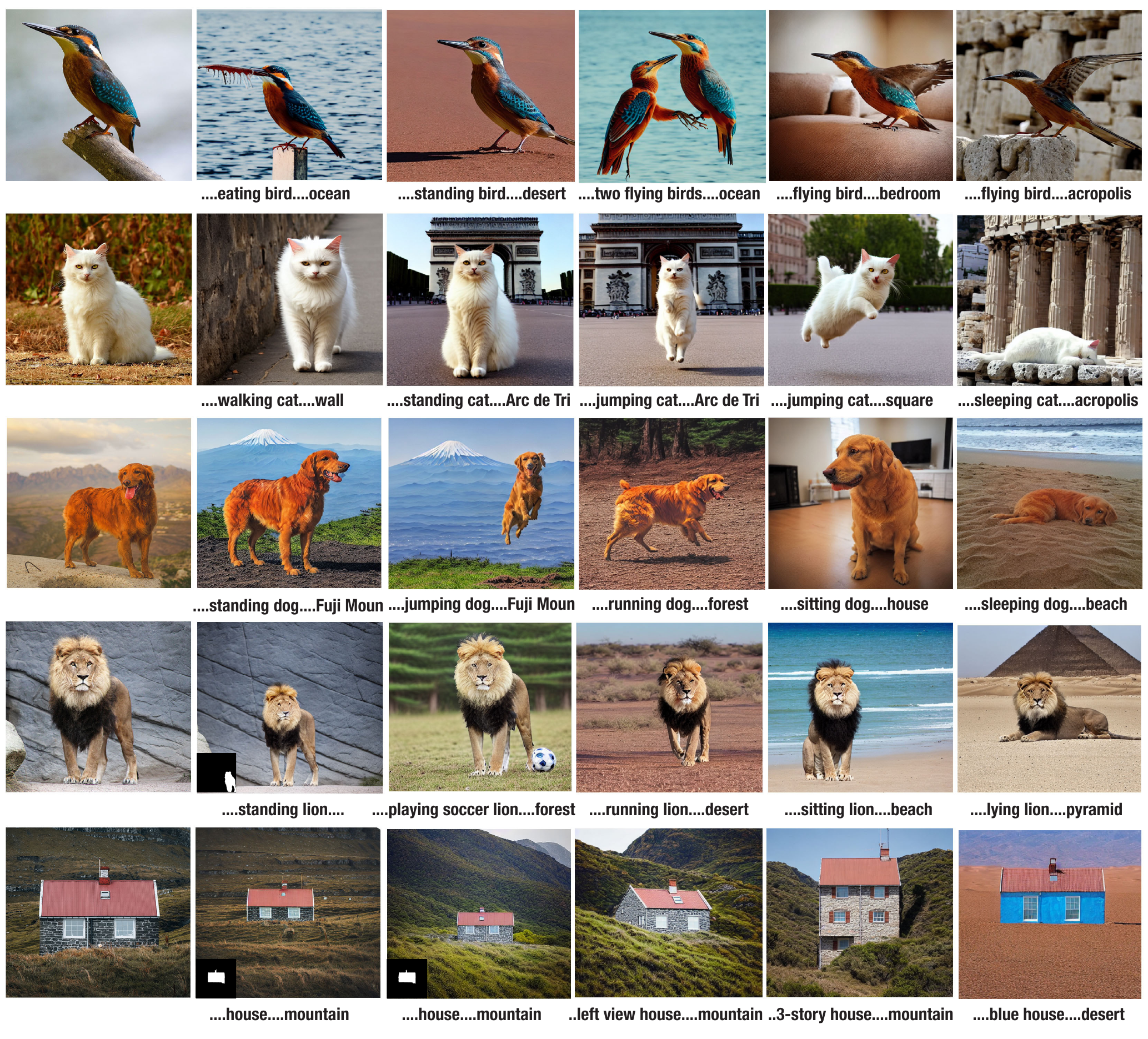}
    \caption{Given a complex text description, the original image (left) is capable of performing multiple editing actions and maintaining similar characteristics of a specific subject. Note that the mask in the bottom left corner is used to change the size of the selected object.}
    \label{fig:visualization}
\end{figure*}

\subsubsection{Model fine-tuing}

We obtain new text embeddings $\boldsymbol {e}_{opt}$ by linearly interpolating multiple optimized text embeddings. Due to the limited number of optimization steps, the resulting embeddings may not lead to a consistent representation of the selected objects or background in the input image. Therefore, we propose a layered diffusion strategy to optimize model parameters while freezing the optimized text embeddings $\boldsymbol {e}_{opt}$. This enables the model to fit the desired image at optimized text embedding points. To achieve the arbitrary modification and combination of foreground object properties and backgrounds, we employ SAM~\cite{kirillov2023sam} to derive $M_{t}$ (object) and $1-M_{t}$ (background) from $O_{t}$ and subsequently obtain $M_{r}$ (object) and $1-M_{r}$ (background) from the reference image $O_{r}$. The aforementioned can be achieved by optimizing the following equations:
\begin{equation} \label{eq_01}
\mathcal{L}_{obj} =  \mathbb{E}_{\boldsymbol{x}_{t}, \boldsymbol{\epsilon} \sim \mathcal{N}(0, \boldsymbol{I})}\left[ \left\| {M}_{t} *(\boldsymbol{\epsilon}- f_{\theta}\left(\boldsymbol{x}_{t}, t, e_{opt})\right)\right\|^{2}\right], 
\end{equation}
\begin{equation} \label{eq_}
\mathcal{L}_{bg} =  \mathbb{E}_{\boldsymbol{x}_{t}, \boldsymbol{\epsilon} \sim \mathcal{N}(0, \boldsymbol{I})}\left[ \left\| (1 - {M}_{r}) *(\boldsymbol{\epsilon}- f_{\theta}\left(\boldsymbol{x}_{t}, t, e_{opt})\right)\right\|^{2}\right], 
\end{equation}

The total loss can be represented as follows:
\begin{equation} \label{eq_2}
\mathcal{L}_{total} = \lambda_{1} \mathcal{L}_{obj} + \lambda_{2}\mathcal{L}_{bg} , 
\end{equation}
This approach enables us to manipulate the foreground object and background independently, allowing for precise control over the final output image.

\subsubsection{Iterative guidance strategy}

We first represent the diffusion process of a pre-trained model as follows:
\begin{equation}\label{e:Phi_series}
\begin{array}{lcr}
     I_T, I_{T-1}, \ldots ,I_0. \ I_{t-1}=D(I_t|y) 
\end{array}
\end{equation}
where $D$ represent an update process: $\mathcal{I} \times \mathcal{C} \rightarrow \mathcal{I}$, $\mathcal{I} \in \mathbb{R}^{H \times W \times C}$ is the image space, and $\mathcal{C}$ is the condition space, and $y \in \mathcal{C}$ is a text prompt. From ${T}$ to ${0}$, $I_{T}$ gradually changes from a Gaussian noise distribution to a desired image by $y$.
Nonetheless, due to the significant gap between the initial image and the desired image in our task, applying the base generative diffusion process with fine-tuned models under condition $y(i.e., e_{opt})$ may still result in failures in modifying object properties in sometimes, such as modifications of actions. 

This issue in image editing is due to the lack of a strong constraint corresponding to the text description of the edited attributes in the diffusion process. The network bias leads the diffusion model to favor object properties in the initial image. To address this, we strengthen the object properties by utilizing the decomposed $\boldsymbol{\hat{e}_{a}}$ in the diffusion process. Specifically, we perform the following approach:
  \begin{equation}\label{e:bootstrap}
 I_{t-1}=\begin{cases}
			D(I_t|\boldsymbol{\hat{e}_{a}}) , & \text{if $t$\%$2=0$}\\
            D(I_t|\boldsymbol{e_{opt}}) , & \text{otherwise}
		 \end{cases}     
  \end{equation}

\subsection{Implementation details}

We adopt the stable diffusion text-to-image model~\cite{rombach2022high} as the baseline for our method. Specifically, we utilize the publicly available v1.4 version of the model, which was pre-trained on the LAION-5B dataset~\cite{schuhmann2022laion} and built upon the latent diffusion model.
We first fine-tune the text embeddings with a learning rate of 1e-3 using Adam~\cite{adam2015} and perform 500 steps in most of our experiments. Subsequently, we fine-tune the diffusion model itself, using a learning rate of 2e-6 and executing 250 steps. We employ an iterative guidance strategy throughout the diffusion process, starting from random noise. This iterative process consist of 50 iterations by default, resulting in more refined results. For one image, it takes about 2 minutes to run on a single NVIDIA A100 GPU.

\section{Experiments}
\subsection{Qualitative Evaluation}

We extensively evaluate our approach using images from various domains and categories. Our method involves a simple text prompt-based editing process, allowing for tasks such as background replacement and object property modification. The images utilized in our experiments are copyright-free on the Internet. We employ a layered editing strategy to ensure robustness and controllability in the editing process. This approach enables multiple editing actions simultaneously on the images, demonstrating excellent editing controllability. The probabilistic diffusion model also motivates us to test our method under different random seeds. By employing our layered diffusion strategy, we can generate images that closely match the provided text descriptions while preserving the critical attributes of the original image in most cases. Our method produces multiple editing results from a single text prompt, providing users with a selection of options to choose from.

In~\cref{fig:visualization}, we present some edited images. 
These images preserve the distinct characteristics of the input image, and they are altered based on text prompts to accommodate a range of editing actions that go beyond mere background replacement and property modification. 
Our method can execute precise editing actions on the images by leveraging reference background or foreground objects. For instance, we can alter foreground objects based on reference foreground object maps or implement background modifications guided by reference background maps. More results can be found in the supplementary material.

\begin{figure*}{}
    \centering
    \includegraphics[width=0.99\textwidth]{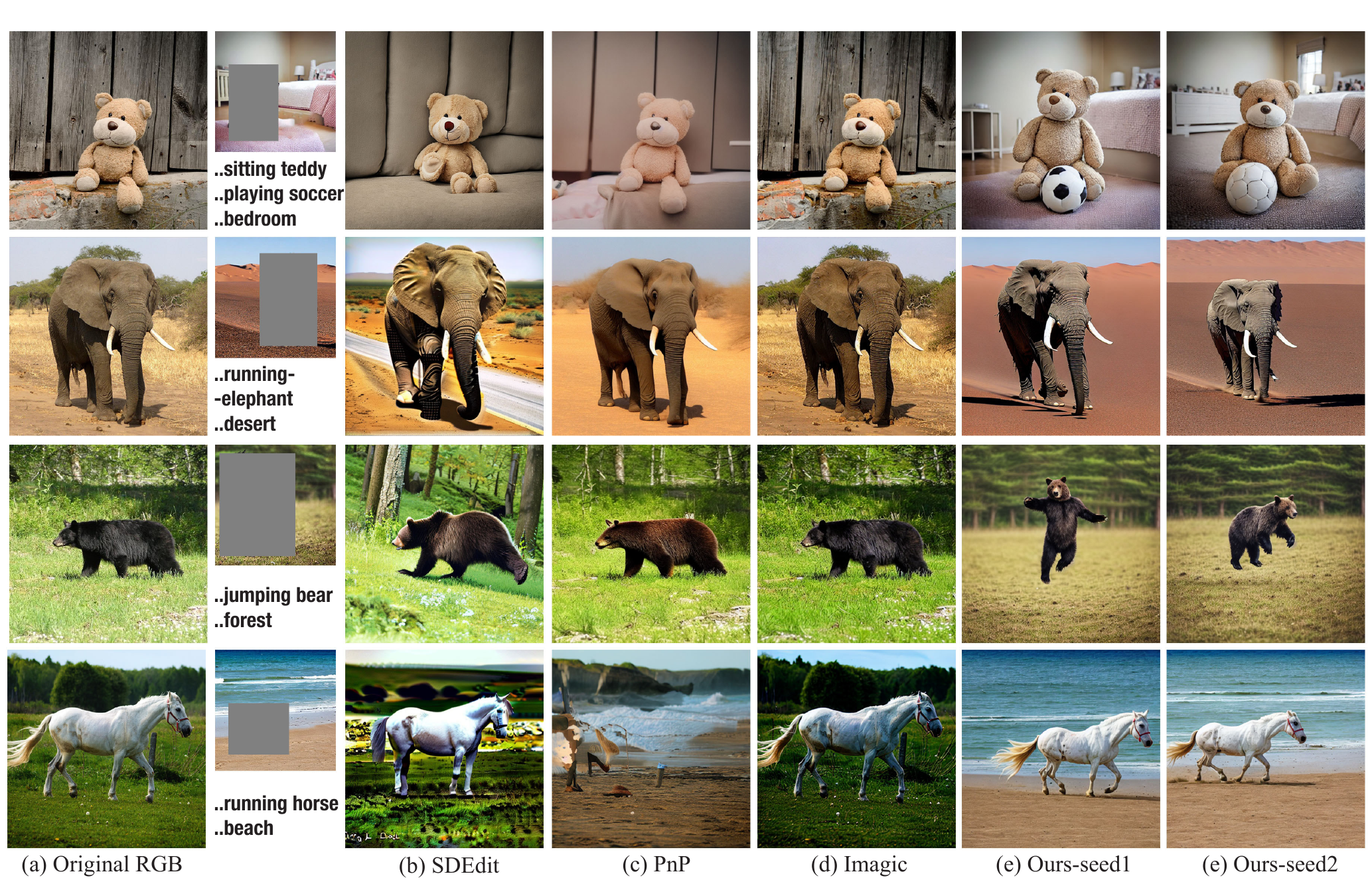}
    \caption{We present several edited images and compare them with similar image editing algorithms, such as SDEdit~\cite{meng2021sdedit}, Imagic~\cite{kawar2022imagic}, and PnP~\cite{tumanyan2022plug}. Our method generates the best results.}
    \label{fig:comparison}
\end{figure*}

\begin{table}[t]
	\centering
	\setlength\tabcolsep{2pt}
		\begin{tabular}{c|cc|c}
			\toprule[1pt]
			\textbf{}& \textbf{Settings}& & \textbf{CLIP score} \\ 
			 & $\mathcal{L}_{obj}$ & $\mathcal{L}_{bg}$   \\ 
			\hline
			(a) & $\times$ & & 0.28 \\
			(b)& &$\times$ & 0.32 \\
			(c)& $\lambda_{1}$ = 1 && 0.29  \\
			(d)& $\lambda_{1}$ = 3&& 0.32\\
			(e)& $\lambda_{1}$ = 1 &$\lambda_{2}$ = 3& 0.28  \\
            (f)&  && \textbf{0.35}  \\
			\bottomrule
	\end{tabular}
\qquad 
		\begin{tabular}{c|cccc|c}
			\toprule[1pt]
			\textbf{}& \textbf{Settings}& &&& \textbf{CLIP score} \\ 
			 & L-c-o & Fine-tune & I-g&$\alpha$  \\ 
			\hline
			(g) & $\times$ & &&& 0.33 \\
			(h)& &$\times$ &&& 0.32 \\
			(i)& & &$\times$&& 0.29  \\
			(j)& & &&= 1& 0.32\\
			(k)& & &&= 0& 0.29  \\
            (f)& & &&& \textbf{0.35}  \\
			\bottomrule
	\end{tabular}
	\caption{Quantitative results with different settings. We report the CLIP score~\cite{hessel2021clipscore} over 300 images.}
	\label{tab:ablation}
\end{table}

\subsection{Comparisons}

We primarily compare our proposed image editing method with previous text prompt methods, such as SDEdit~\cite{meng2021sdedit}, Imagic~\cite{kawar2022imagic}, and PnP~\cite{tumanyan2022plug}. It is worth noting that Imagic~\cite{kawar2022imagic} necessitates fine-tuning of both the network and text embeddings, while our method adopts a similar fine-tuning approach.

As shown in~\cref{fig:comparison}, non-rigid editings, such as jumping and rotation, have significant challenges in image editing tasks. This complexity leads to the failure of both PnP~\cite{tumanyan2022plug} and SDEdit~\cite{meng2021sdedit} in performing editing actions. Additionally, Imagic~\cite{kawar2022imagic} tends to produce overfitting of the original image and text embeddings during training, thereby making accurate image editing difficult, especially when modifying text prompts go beyond attribute editing and involve adding other editing actions, such as foreground-background editing simultaneously. In contrast, our approach adopts a layered strategy that allows for the simultaneous execution of multiple editing actions. As a result, our method achieves impressive results in real image editing tasks. The last two columns of~\cref{fig:comparison} show the edited results generated by employing different random seeds. Our method outperforms others in multitask editing performance.

In~\cref{fig:comparison}, we generate a reference background image from the diffusion model~\cite{rombach2022high}, and our layered diffusion approach allows us to make the edited image as close as possible to the reference background image. We can also choose our reference image as long as it is close to the perspective of the original image. We show more results in the supplementary material.

\begin{figure*}{}
    \centering
    \includegraphics[width=0.99\textwidth]{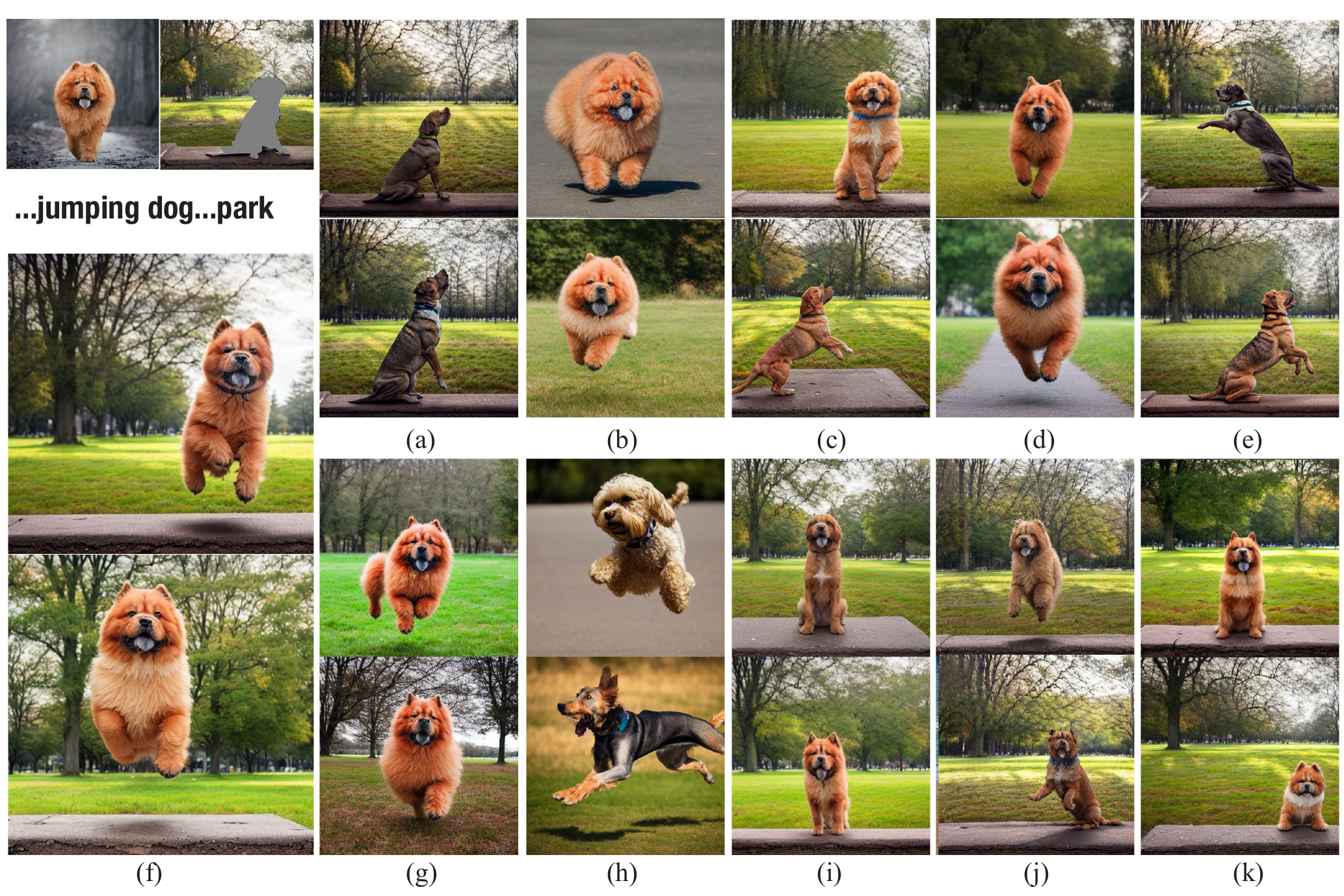}

    \caption{We present the edited images with different settings. For each setting, we show two generated images using different random seeds. (f) illustrates the final edited results.}
    \label{fig:ablation}
    \vspace{-1em}
\end{figure*}

Text-based image editing methods are a relatively new direction, and there is currently no standard benchmark for evaluating our approach. Although Imagic~\cite{kawar2022imagic} propose the \textit{TEdBench}, it includes only a single non-rigid edit, which is also not fully applicable to our approach.
To further assess the quality of our generated results, we utilize the \textit{TEdBench} dataset to generate over 300 images per method for a preliminary evaluation. The supplemental material includes the used text prompts. 
We use a CLIP-based text-image clip score~\cite{hessel2021clipscore,radford2021learning}, which measures the cosine similarity between the text prompt and the image embeddings. As our method aims to maintain proximity between selected object features before and after non-rigid editing, the CLIP score does not effectively demonstrate the superiority of our approach (see~\cref{fig:ablation} (f)-(g) and~\cref{tab:ablation} (f)-(g)). However, it still partially reflects the state of our image editing, such as changes in the motion of the selected object. ~\cref{tab:ablation} provides an approximate representation of the CLIP scores for several methods, and our approach achieves the highest score.

\subsection{Ablation Study}

In this section, we present a comprehensive analysis of the three modules employed in our method. We utilize the \textit{TEdBench}~\cite{kawar2022imagic} dataset and generate over 300 images using 20 different random seeds. As an auxiliary objective evaluation metric, we employ the text-image CLIP score~\cite{hessel2021clipscore}, which is presented in~\cref{fig:user}.
Furthermore, we present the specific performance of each component in~\cref{tab:ablation}. As mentioned previously, the CLIP score may not fully capture the suitability of our method as it primarily focuses on the alignment between images and text. For instance, the results of (b), (g), and (i) show high CLIP scores, but their object features significantly differ from the reference images. 

As shown in~\cref{fig:ablation} and~\cref{tab:ablation}, (a) does not utilize $\mathcal{L}_{obj}$, resulting in a background that matches the reference image, while the properties of the foreground objects differ substantially. On the other hand, (b) demonstrates that $\mathcal{L}_{bg}$ preserves a more similar background. (c), (d), and (e) analyze the impact of different weights assigned to the two losses, which affect the similarity of the background and foreground objects. In this paper, we mostly set $\lambda_{1}$ to 2 and $\lambda_{2}$ to 1, except when $\lambda_{1}$ is set to 3 for smaller foreground objects.
(g), (h), and (i) validate the effectiveness of each of the three modules in our method. (g) enhances the similarity of the background, (h) controls the global features, and (i) significantly increases the percentage of image generation results that satisfy the description text, rising from 43$\%$ to 81$\%$.

\begin{figure*}{}
    \centering
    \includegraphics[width=0.99\textwidth]{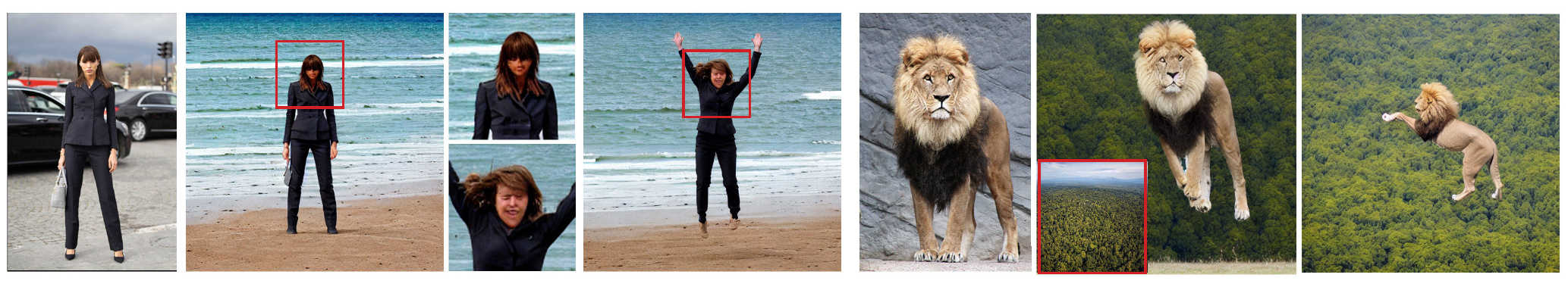}
    \caption{We present several failure cases, including artifacts on faces and significant disparities in the camera angles of the images.}
    \label{fig:fail}
\end{figure*}

\subsection{User Study}

\begin{figure*}{}
    \centering
    \includegraphics[width=0.6\textwidth]{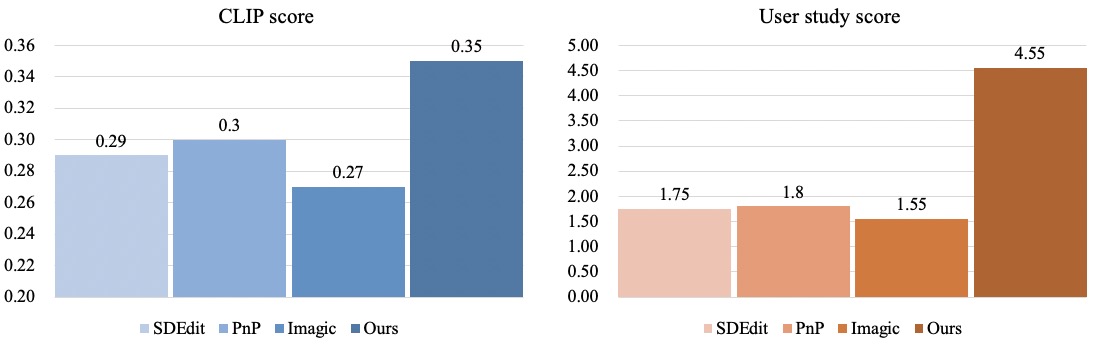}

    \caption{We compare several image editing methods using the CLIP and subjective user perception scores. Our method achieves a relatively higher score.}
    \label{fig:user}
\end{figure*}

Furthermore, we conduct a user study to evaluate and compare the subjective perception of our method with several other approaches. To ensure a fair comparison, we randomly select ten generated images and utilize two random seeds to generate our results. We present two discriminant conditions for each image: background similarity and action similarity. We then ask 20 participants to rate the resulting images on a scale ranging from 1 to 5, with a rating of 5 indicating a very good match and 1 indicating a very poor match. The histogram on the right-hand side of ~\cref{fig:user} shows the average scores. Remarkably, our method achieves optimal subjective performance compared to the other methods.

\section{Limitations}

While our approach demonstrates superior performance in achieving controlled image editing and accomplishes remarkable results in tasks involving multiple editing actions, it is essential to acknowledge three significant challenges. (1) Dealing with fine-grained tasks is still challenging for our method while we rely on a pre-trained text-to-image diffusion model and the problem of overfitting that occurs during model fine-tuning. \cref{fig:fail} demonstrates that our method will produce artifacts when confronted with textures with intricate details or facial features.
(2) As shown in the~\cref{fig:fail}, another challenge arises when there is a notable disparity in camera angles between the input reference image and the desired edited image, leading to the creation of visually inconsistent scenes. This limitation can be mitigated by incorporating additional descriptions about the camera position in the target text.
(3) We need to fine-tune the model to accommodate the reference image. Appropriately fine-tuning specific parameters may be required for unconventional or atypical manifestations to generate good results in sometimes.

\section{Conclusion}

We propose \textit{LayerDiffusion}, a semantic-based layered image editing method that simultaneously edits specific subjects and backgrounds using a single input image. \textit{LayerDiffusion} preserves the unique characteristics of the subjects while integrating them seamlessly into new scenes. Extensive experimentation demonstrates that our method generates images closely resembling the feature of the input images, surpassing existing approaches in editing quality and controllability. User studies confirm the subjective perception of the generated images, aligning with human expectations.
Our contributions include introducing \textit{LayerDiffusion} as the first method for simultaneous editing of specific subjects and backgrounds. We develop a layered diffusion training framework for controllable image editing, which opens up new possibilities for text-guided image editing tasks. We may focus on preserving complex textures and facial features in the future.

\bibliographystyle{ieee}
\bibliography{neurips}

{
\small



}

\appendix
\newpage

\begin{figure*}[!h]
    \centering
    \includegraphics[width=0.8\textwidth]{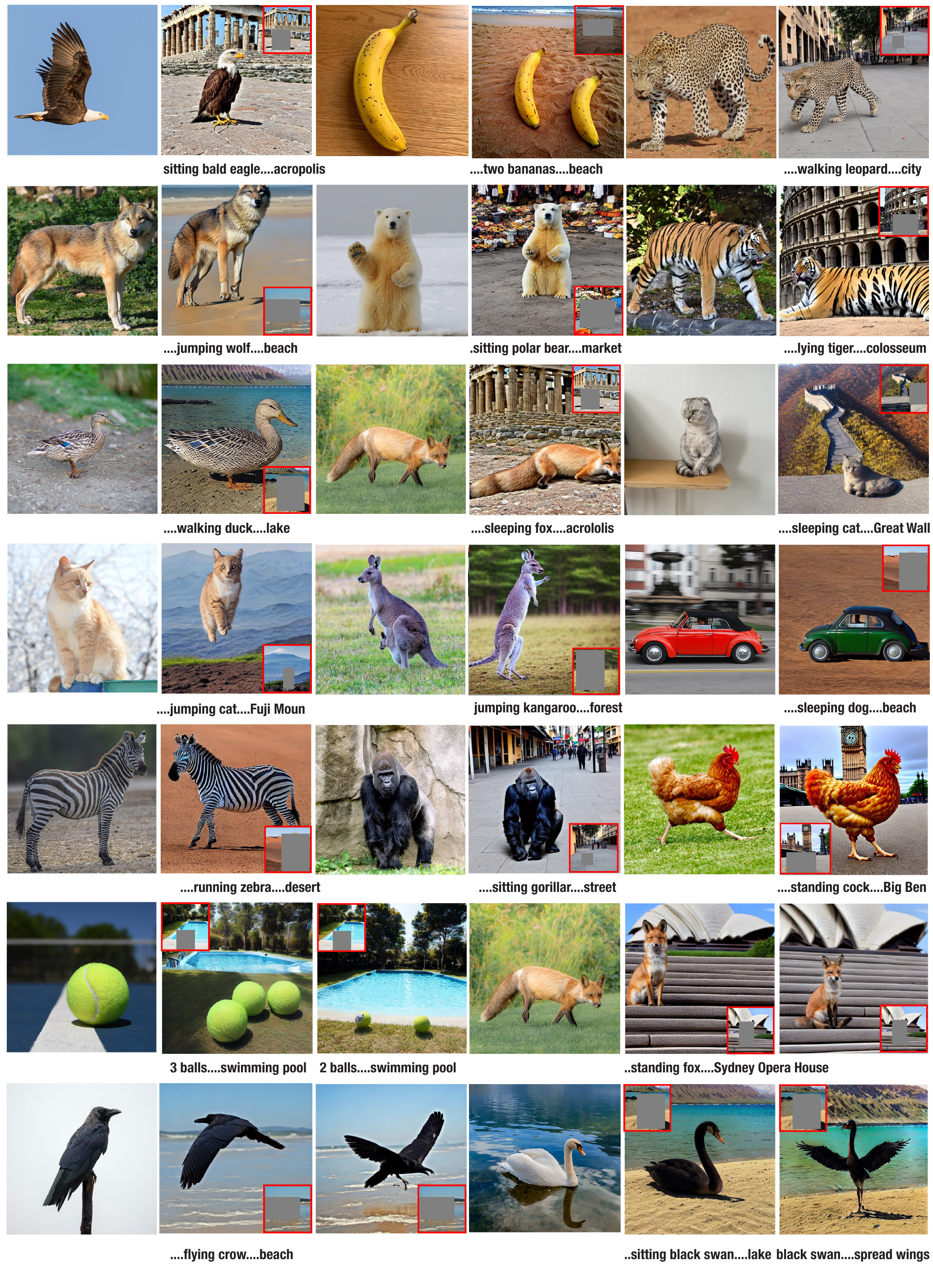}
  
    \caption{We present more edited results. Each triplet consists of the original image on the left, the edited result on the right, and a small reference image enclosed within a red box.}
    \label{fig:fail1}
\end{figure*}
\vspace{-2em}
\section{More Results}
We present additional image editing results in Figure 1, demonstrating the versatility of our approach across various image types.

\section{More Ablation Study}

\subsection{Training}

We conduct an analysis of the number of layered controlled text embeddings optimization steps, as well as the fine-tuning steps. When the number of optimization steps for text embeddings is too low, the similarity of image features is relatively low. Conversely, when the number of steps is excessively high, the quality of generated images deteriorates. Similarly, when the number of fine-tuning steps is insufficient, the generated images exhibit lower feature similarity. Conversely, with an excessively high number of fine-tuning steps, the content of the images gradually deviates from the textual cues. After extensive experimentation, we determine that the optimal configuration for the number of layered controlled optimization steps is 500, while the number of fine-tuning steps is set to 250.

\begin{figure*}[!h]
    \centering
    \includegraphics[width=0.99\textwidth]{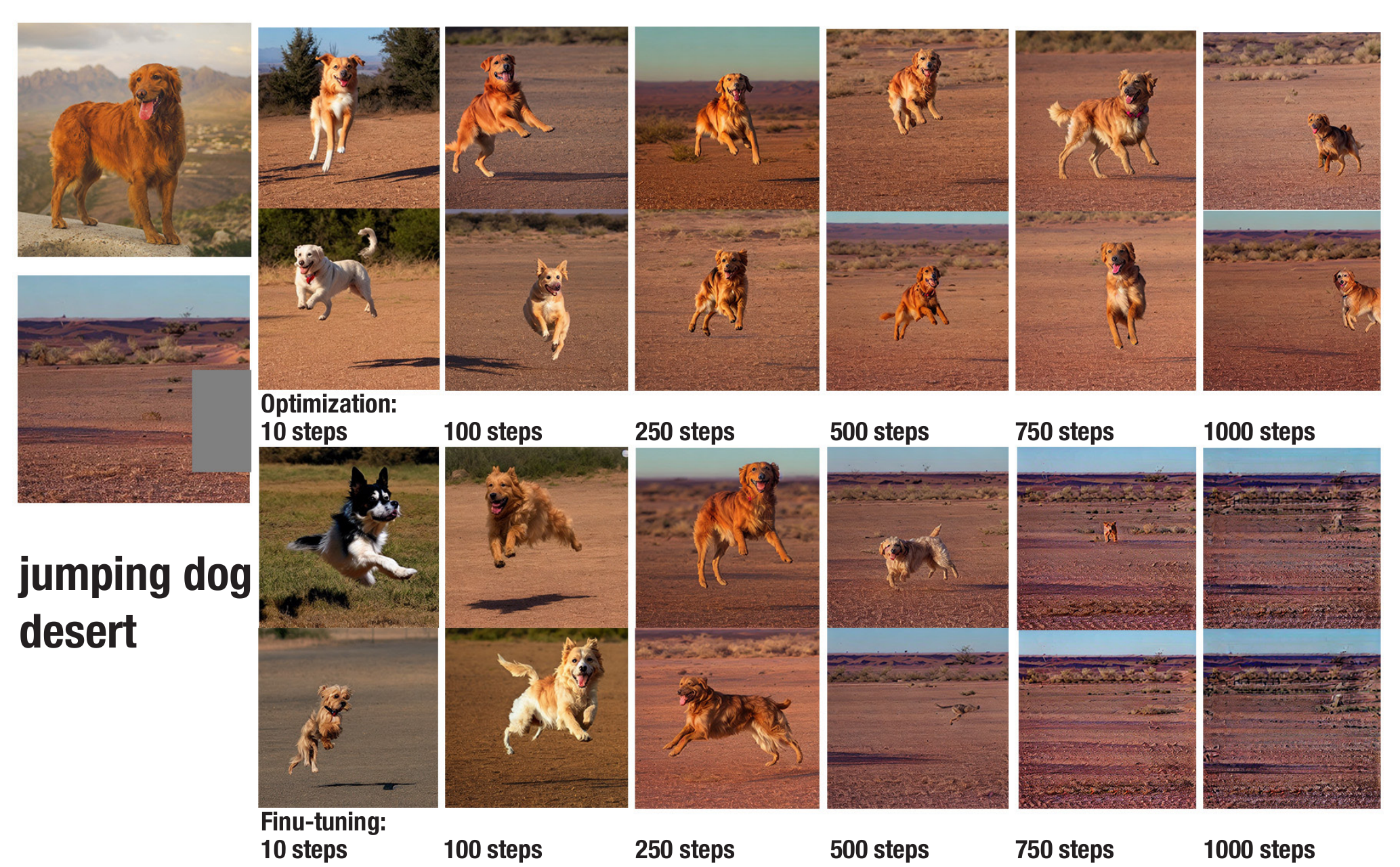}

    \caption{We employ two random seeds to show the image editing results as the number of steps gradually increase.}
    \label{fig:trainstep}
\end{figure*}

\subsection{Text embeddings interpolation}
\begin{figure*}[!t]
    \centering
    \includegraphics[width=0.99\textwidth]{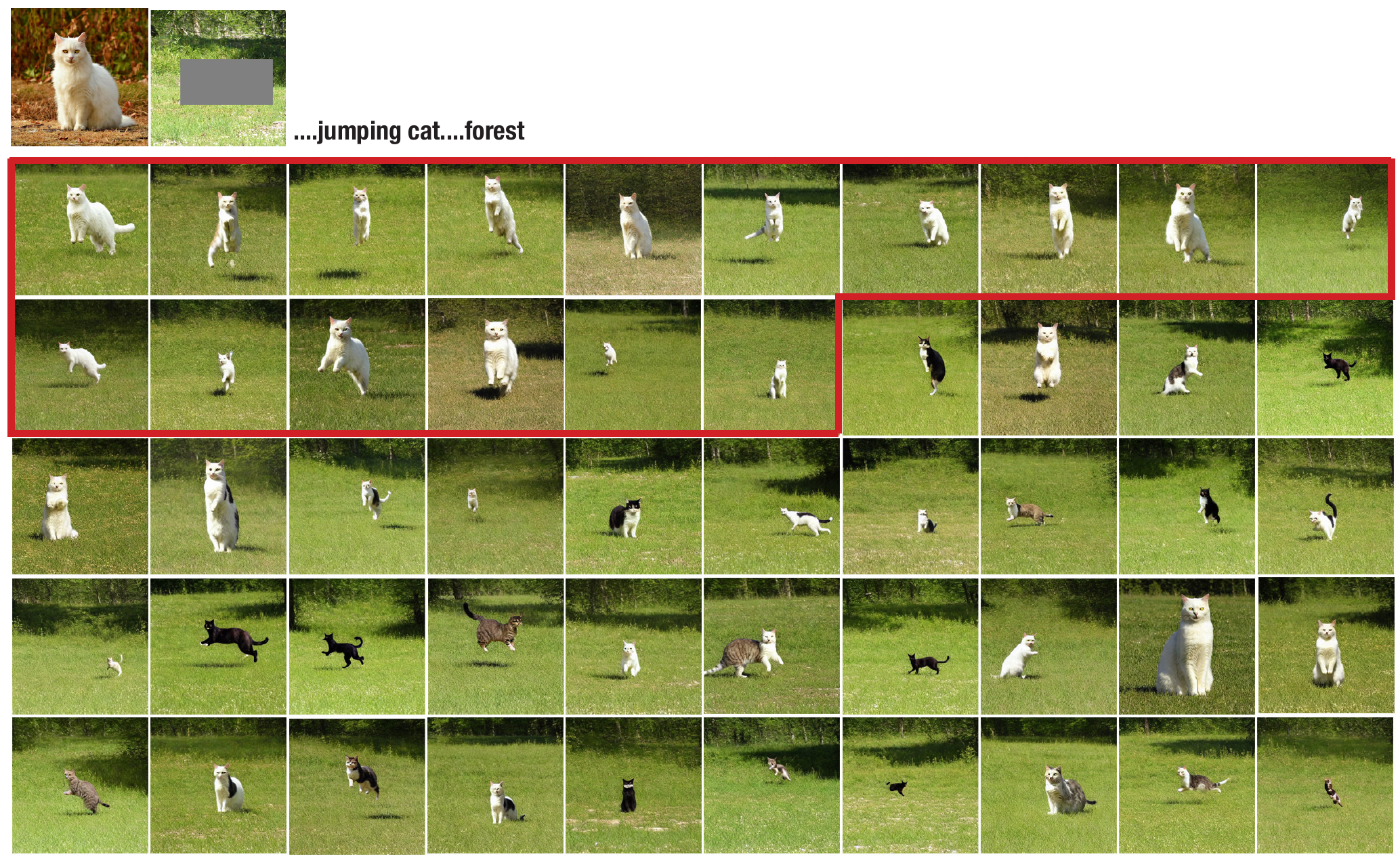}
    \caption{We show the results of 50 randomly generated edited image with $\alpha$ set to 1, where the red boxes indicate the images that satisfy the text prompt.}
    \label{fig:10}
\end{figure*}

\begin{figure*}[!t]
    \centering
    \includegraphics[width=0.99\textwidth]{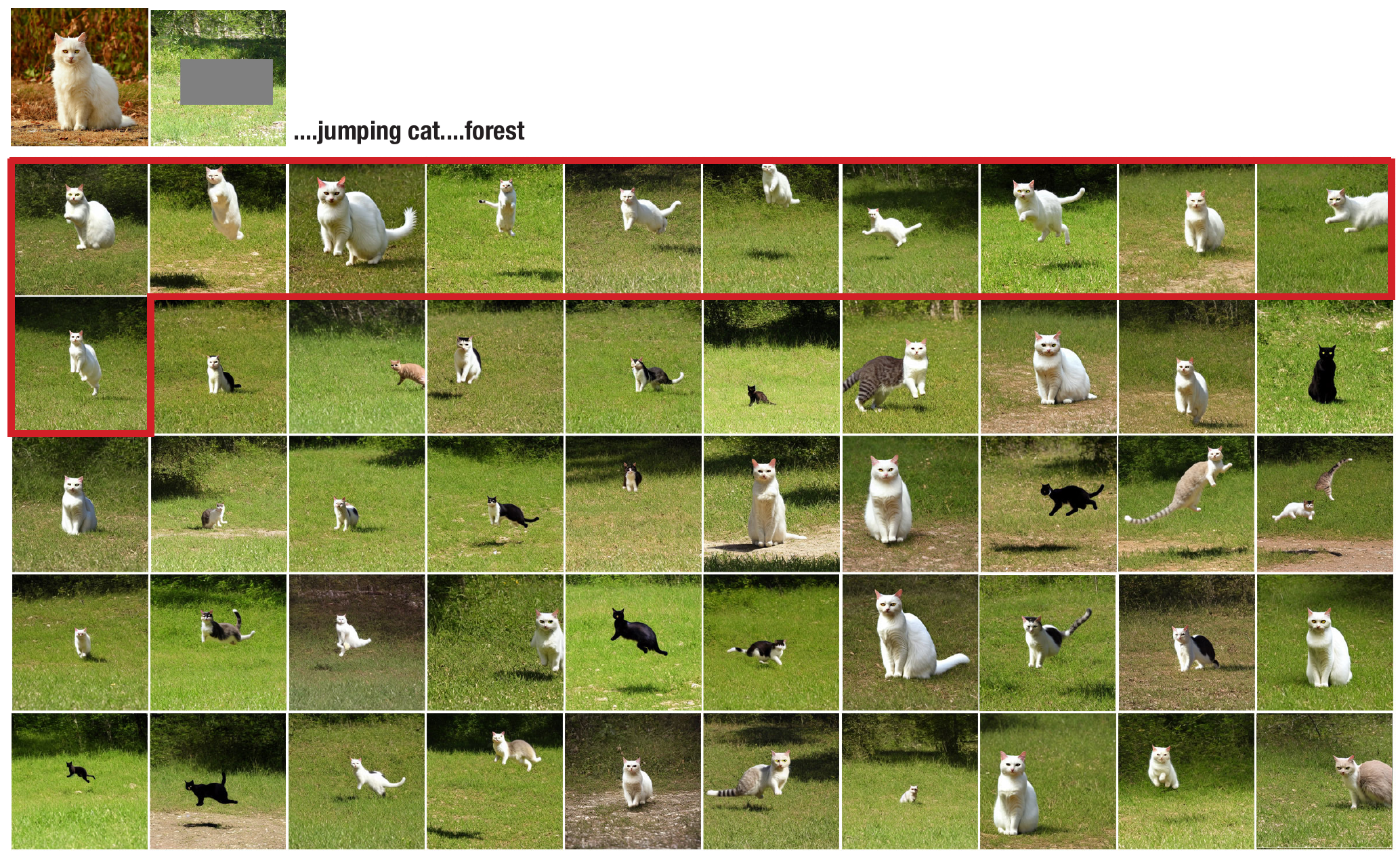}
    \caption{We show the results of 50 randomly generated edited imagewith $\alpha$ set to 0, where the red boxes indicate the images that satisfy the text prompt.}
    \label{fig:01}
\end{figure*}

\begin{figure*}[!t]
    \centering
    \includegraphics[width=0.99\textwidth]{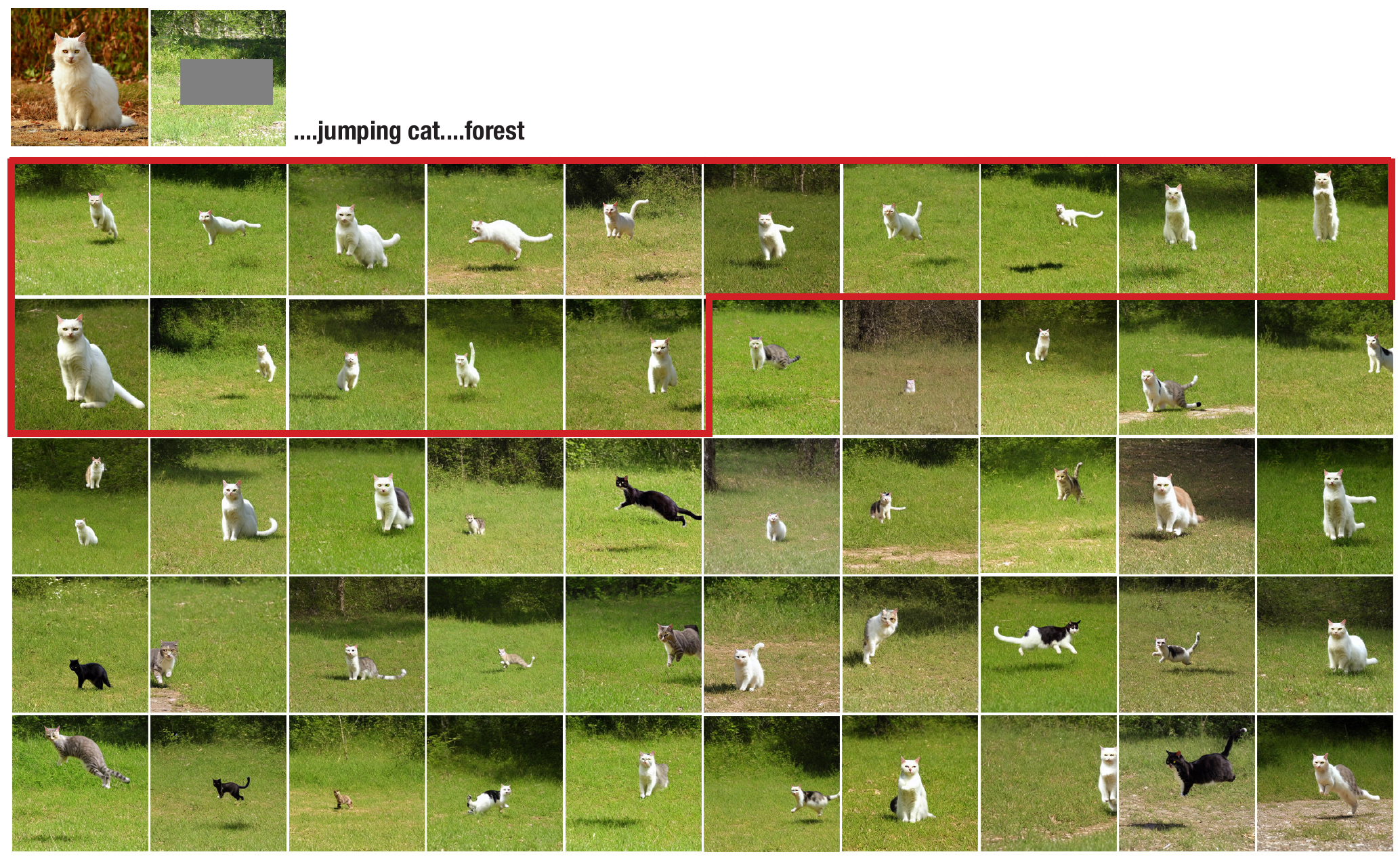}
    \caption{We show the results of 50 randomly generated edited image with $\alpha$ set to 0.3, where the red boxes indicate the images that satisfy the text prompt.}
    \label{fig:73}
\end{figure*}

We first review the process of linear interpolation of the optimized text embedding in the main paper:
\begin{equation}
\boldsymbol {e}_{opt} = \alpha * \boldsymbol {\hat{e}}_{a} + (1 - \alpha) * \boldsymbol {\hat{e}}_{b},
\end{equation}
This linear interpolation is analyzed a bit in Imagic~\cite{kawar2022imagic}. But our approach differs from it in that we are interpolating two optimized text embeddings while preserving the features of both text embeddings. Next, we will analyze the choice of the value of $\alpha$. In \cref{fig:01} and \cref{fig:10} , we set $\alpha$ to 1 and 0. Fifty different sets of results are generated using different random seeds under the same training parameters, and the image editing results that satisfy the text prompt are less. In \cref{fig:73}, we also set $\alpha$ to 0.3, which is the opposite of our setting in the experiment, and the image editing results that satisfy the text prompt are also less. This indicates that an appropriate value of a can improve the image editing results that satisfy the text prompt.

\begin{figure*}[!t]
    \centering
    \includegraphics[width=0.99\textwidth]{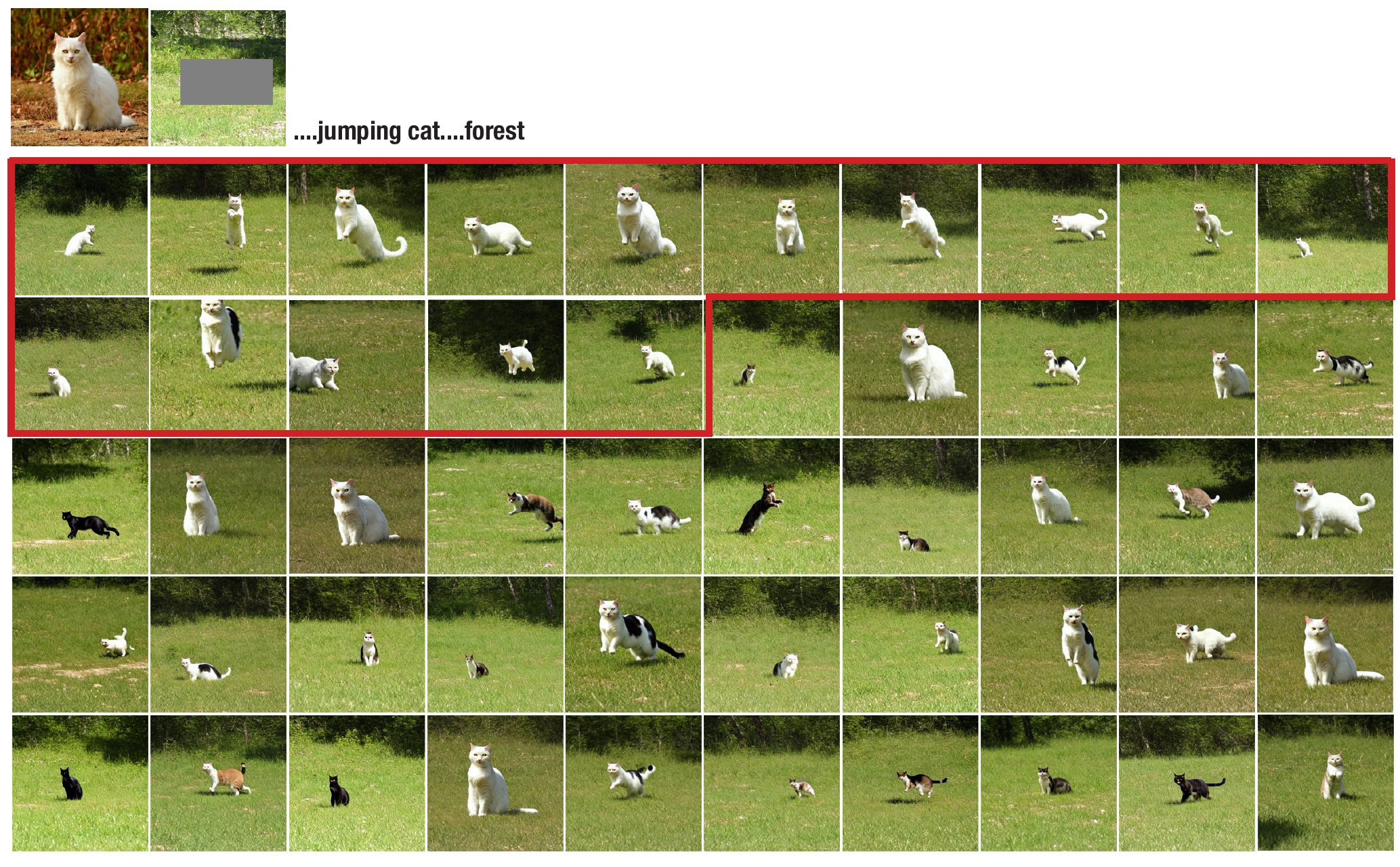}
    \caption{We present 50 sets of randomly generated results without employing an iterative guidance strategy. The red boxes indicate the edited images that satisfy the text prompt.}
    \label{fig:noig}
\end{figure*}

\begin{figure*}[!t]
    \centering
    \includegraphics[width=0.99\textwidth]{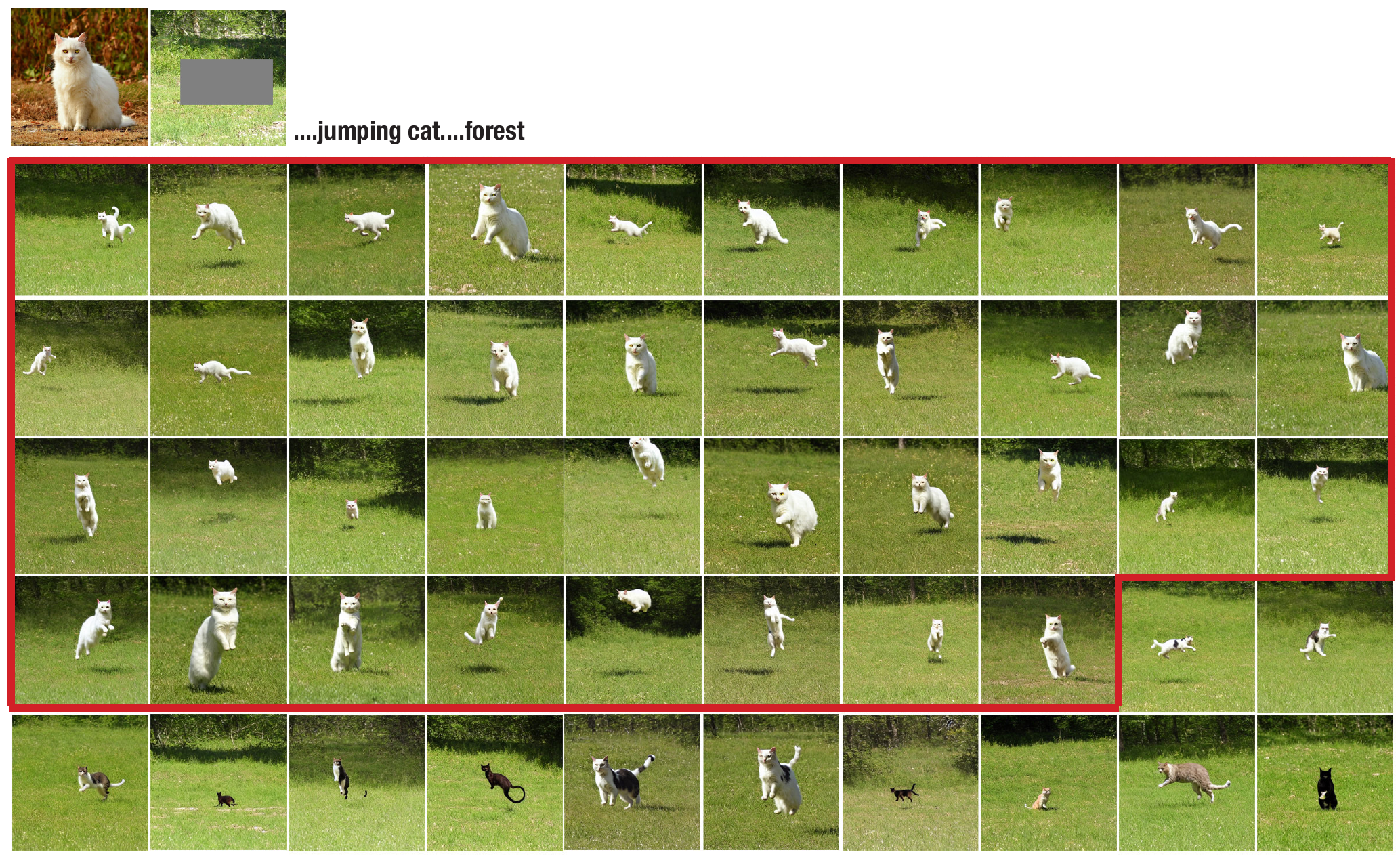}
    \caption{In our final experiment, we include all the utilized modules and parameters, resulting in a collection of image editing outputs. The number of edited images satisfying the text prompt within the red boxes is remarkably large.}
    \label{fig:our}
\end{figure*}

We add a strong constraint on the object properties, which also substantially increases the validity of our generated results. In \cref{fig:noig}, we remove this module and the number of valid editing results is drastically reduced compared to our final results in \cref{fig:our}.

\end{document}